\documentclass{llncs}
\usepackage{amsmath,amssymb}
\usepackage{graphicx}
\usepackage{cite}
\usepackage{url}
\usepackage{booktabs}

\usepackage{booktabs}
\usepackage{multirow}
\usepackage{array}
\usepackage{amsmath}
\usepackage{tabularx}
\usepackage{csquotes}
\usepackage{listings}
\usepackage{ulem}

\usepackage{color}

 \newcolumntype{b}{>{\hsize=1.5\hsize}X}
\newcolumntype{s}{>{\hsize=.45\hsize}X}
\newcolumntype{m}{>{\hsize=.9\hsize}X}

\begin{document}

\title{Learning Representations for Soft Skill Matching}

\author{%
    Luiza Sayfullina,
    Eric Malmi  \and Juho Kannala 
}


 \institute{
     Aalto University, Finland \\
}

\maketitle
\begin{abstract}
    Employers actively look for talents having not only specific hard skills but also various soft skills. To analyze the soft skill demands on the job market, it is important to be able to detect soft skill phrases from job advertisements automatically. However, a naive matching of soft skill phrases can lead to false positive matches when a soft skill phrase, such as \texttt{friendly}, is used to describe a company, a team, or another entity, rather than a desired candidate. \\
	In this paper, we propose a phrase-matching-based approach which differentiates between soft skill phrases referring to a candidate vs. something else. The disambiguation is formulated as a binary text classification problem where the prediction is made for the potential soft skill based on the context where it occurs. To inform the model about the soft skill for which the prediction is made, we develop several approaches, including soft skill masking and soft skill tagging.\\ 
We compare several neural network based approaches, including CNN, LSTM and Hierarchical Attention Model. The proposed tagging-based input representation using LSTM achieved the highest recall of 83.92\% on the job dataset when fixing a precision to 95\%. 
      
\end{abstract}
\section{Introduction}

Since 1980, the number of jobs with high social skills requirements has increased by around 10\% from the total US labor force\cite{future2030}. When an industry becomes more competitive, job applicants need to possess additional skills that help them to start smoothly at work and to be distinguished from candidates with similar qualifications
\cite{schulz2008importance}.

According to Collins dictionary, soft skills are desirable qualities for certain forms of employment that do not depend on acquired knowledge: they include common sense, the ability to deal with people, and a positive flexible attitude. 


Detecting soft skills in the unstructured text corpora is not straightforward, since some matches can turn out to be false positives. For example, the word \texttt{tolerant} might be matched as a part of \texttt{fault tolerant systems} phrase. So, the task is to increase the precision of soft skill detection with an additional requirement: to match only those soft skills that would describe a candidate. For example, \texttt{friendly team} would be a false positive, despite \texttt{friendly} is a soft skill.

The task of detecting candidate soft skills is related to word sense disambiguation (WSD), where the sense of a word being used in a particular context has to be determined \cite{jurafsky2009speech}. However, our task is different since we are not aiming at mapping a skill to a particular sense such as a one from WordNet\cite{Miller:1995}. Instead, the goal is to differentiate between the two cases of interest without requiring exact sense. Besides, we work both on the word and phrase levels.

Dependency parsing \cite{nivre2003efficient} could be helpful for disambiguation of the skill by detecting the entity to which the skill refers. However, the amount of possible entities referring to the candidate is quite high, since among such obvious entities as \texttt{candidate}, \texttt{individual}, a candidate might be referred by a profession name or a university degree level. Thus, entities referring to candidate and others can be hard to differentiate. Working with skill phrases, also makes it harder to use dependency parsing, that usually outputs a relation between a pair of words.

Massive skill extraction was done and actively employed by LinkedIn \cite{bastian2014linkedin}, although with a focus on hard skills. They observed that people provide a list of comma separated skills in the specialties section of their profile that can be used for mining potential skills. The authors disambiguated skills by using clustering based on the co-occurring phrases. Some skills were tagged with many senses belonging to different clusters with an industry label. Kivim\"{a}ki et al. \cite{kivimaki2013graph} suggested a system for automatic detection of new skills in free written text using spreading activation algorithm on the Wikipedia graph and skill lists from LinkedIn. However, these skill extraction approaches aim at mining and extracting new skills, rather than using a ready list of skills and detecting them.

We formulate the problem of soft skill matching as a binary classification, where the positive class refers to the skill describing the candidate and the negative class refers to all other cases. We do classification on the text snippets or the context sentences where the soft skill occurs. 

One baseline approach for tackling the classification task is to use any existing text classifier to classify the context sentence as such. A limitation of this approach is that the classifier is not explicitly informed about the skill it is trying to disambiguate. To address this limitation, we investigate various alternative input representations where the context sentence is modified in order to indicate the soft skill to be classified. 



The contributions of this work can be summarized as follows:
\begin{enumerate}
\item We are the first one to formalize the problem of soft skill matching and release a benchmark dataset collected via crowdsourcing for this problem.
\item We introduce various input representations for the classifier, including modifying the context with soft skill masking, soft skill tagging and augmenting it with soft skill embeddings.
\item We evaluate the performance of several strong text classification baselines, including LSTM, CNN and the Hierarchical Attention Model, using the proposed input representations. Our experiments show that the best performance is achieved using soft skill tagging as well as unmodified sentence representations.
\end{enumerate}


The paper is organized as follows. First, we provide a motivation for our work by analysis of entities that soft skills relate to in Section~\ref{analysis}. In Section~\ref{cv_job_datasets}, we briefly describe both CV (Curriculum Vitae) dataset and the job ads dataset. In Section~\ref{crowdsourcing}, the process of obtaining the data with the help of crowdsourcing is explained. Then, in Section~\ref{methods}, we give a brief overview of applied neural network--based classifiers and describe the proposed approach for input representations. Experimental setup and the analysis of the obtained results with different approaches is presented is Section~\ref{results}.

\section{Motivation}\label{analysis}

\begin{figure}
\centering
\begin{minipage}{.45\linewidth}
  \includegraphics[width=\linewidth]{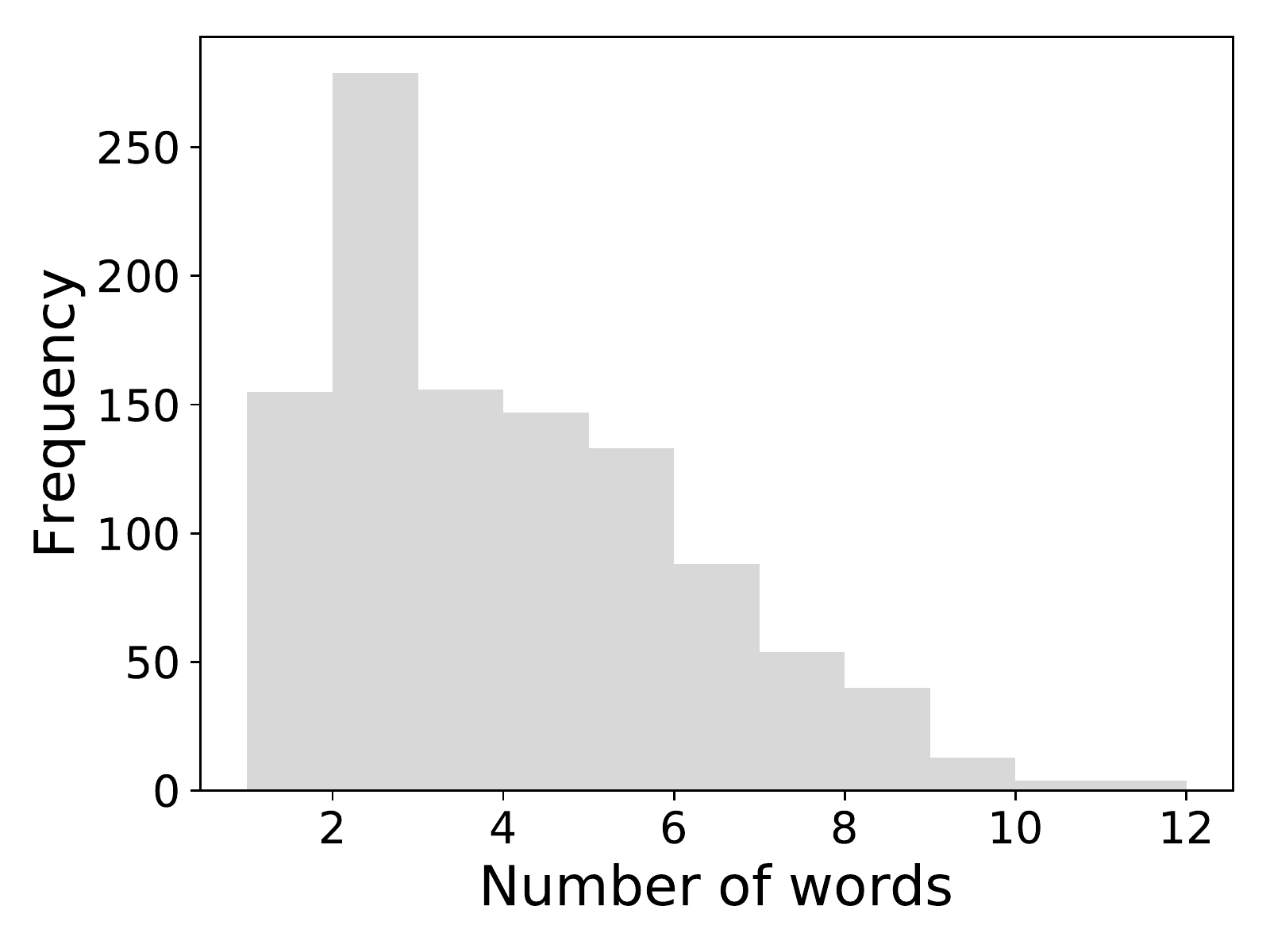}
  \caption{The histogram of soft skill phrase lengths. Due to presence of stop words, the length reaches 12 words. The majority of the skills are shorter than 6 words.}
  \label{lengths}
\end{minipage}
\hspace{.01\linewidth}
\begin{minipage}{.45\linewidth}
  \includegraphics[width=\linewidth]{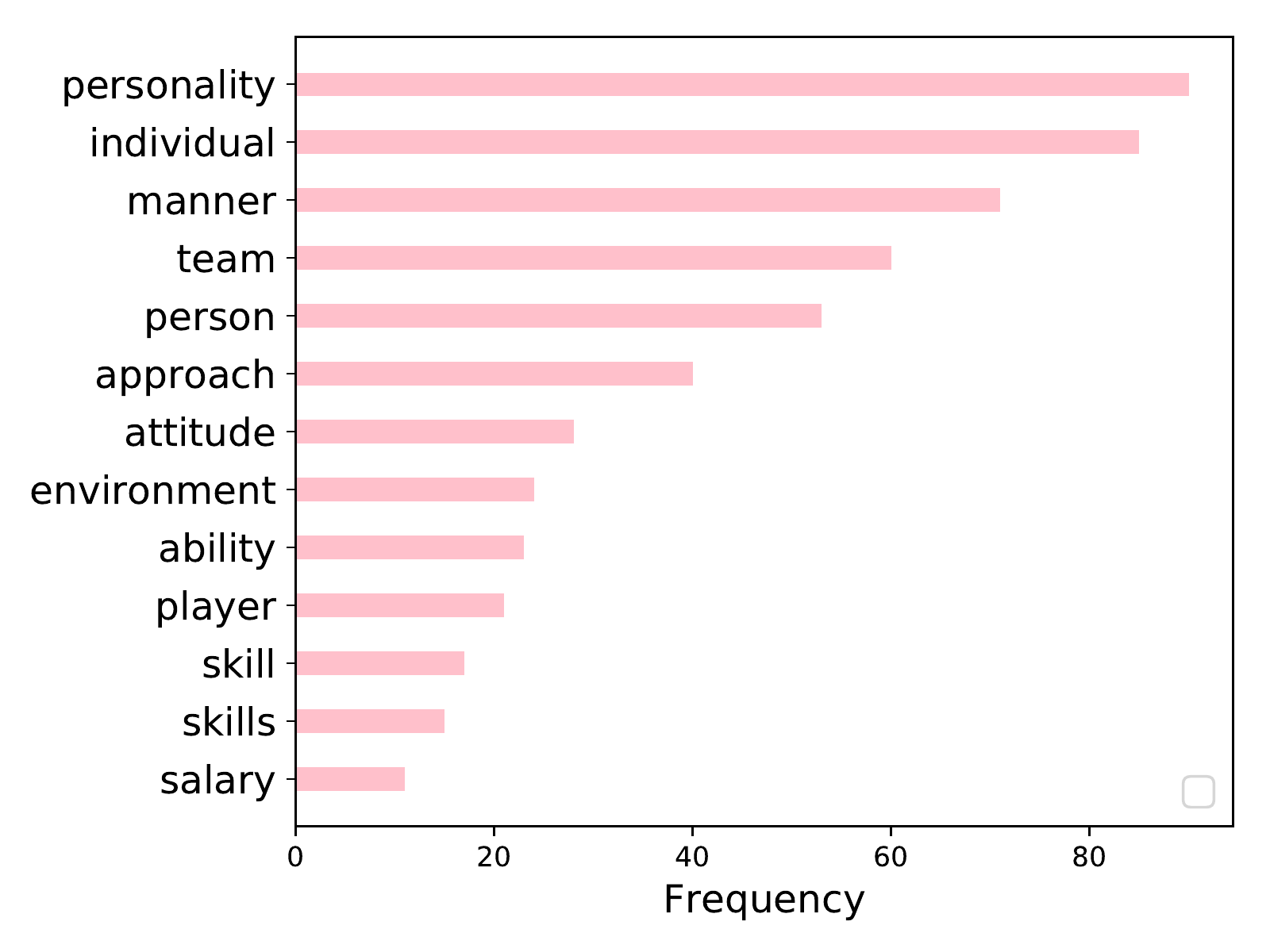}
  \caption{Frequencies of the most common nouns used with one-word adjective soft skills. The statistics was obtained using 10,000 sampled sentences from the job ads dataset.}
  \label{top_entities}
\end{minipage}
\end{figure}


When it comes to detection, some skills can refer to various entities and not only to a candidate. Consider the following two sentences found from job ads: ``Produce \texttt{accurate}, legible documentation packs...'' and ``Fault \texttt{tolerant} systems...''. In these cases, words \texttt{accurate} and \texttt{tolerant}---two potential soft skills---do not explicitly refer to the qualities of the desired candidate. In general, shorter soft skill phrases, especially one-word skills, are more probable to result in false positive matches.
Some skills like \texttt{responsible} can be treated alone as soft skill, but with additional phrase, like \texttt{responsible for finding new clients}, these become job duties.

In order to understand the problem better, we use Stanford Dependency Parser~\cite{chen2014fast} to find entities with adjectival modifier (amod), particularly \texttt{adjective + noun} phrases. Since we are interested in cases where certain adjectives or skills are used to describe other entities, this method allows us to mine easily some of such cases. Thus, we selected only \texttt{amod} relations with adjectives that are part of soft skill list from the first 10,0000 sentences from the job ad dataset (to be described in the next section) for the preliminary study.

The top nouns used with one-word soft skills are shown in Figure~\ref{top_entities}. The majority of these nouns, like \texttt{personality}, \texttt{individual}, \texttt{person} indeed refer to the candidate, but frequent non-candidate nouns like \texttt{team}, \texttt{environment}, \texttt{salary} can be found as well. 

Despite the fact that dependency parsing helps to find some of the entities that soft skills describe, this method occasionally fails to find the objects to which a soft skill refers. This can happen when objects are simply absent in the sentence, e.g. \texttt{communication skills are required at work}. Secondly, dependency parsing usually finds the relations for a pair of words, that makes it hard to apply for soft skill phrases.





\section{Datasets}\label{cv_job_datasets}

In this work we use three datasets for the analysis: the first is a corpus of job advertisements (ads), the second is a corpus of CVs or resumes of people and the third is a resource with soft skills list. Both job ads and CVs contain soft skills and candidate descriptions. However, job ads provide more examples with cases where soft skills are used in different contexts and describe company, team, environment, etc. The purpose of using a CV dataset is to have additional data from a different domain to test our approach. Both datasets are used for soft skill annotation that provides data for using supervised approaches. Finally, the third dataset is the set of soft skills that we learn to disambiguate. 


We got access to a comprehensive list of soft skills. 
There are 1072 skills and 234 skill clusters respectively. Clusters group skills with a similar meaning but phrased differently. For example, the \texttt{polite} cluster consists of 
\texttt{polite manners}, \texttt{polite} and
\texttt{ability to speak politely}. 
155 skills out of 1072 are one-word, while the longest one consist of 12 words. The histogram of soft skill phrase lengths is shown on Figure~\ref{lengths}. We see that most of the soft skill phrases consist of less than 6 words.


\subsection{Kaggle job ads and CV corpora}

In order to obtain job postings, we have used a publicly available dataset from Kaggle with UK job ads.\footnote{The dataset is available at: \url{https://www.kaggle.com/c/job-salary-prediction}} It contains around 245,000 job entries for training, each of them having a job description paragraph. Each job description consists of several sentences, providing such information as the desired candidate qualities, work duties, requirements and background information about the company or team to work with.

In Kaggle this dataset is used for salary prediction task and therefore contains salary information as well as job title, location, contract type, company name and job category. We use only ``Train\_rev1'' file since it contains a sufficient amount of data for our analyses. 




We collected manually 525 CV samples from \textit{indeed.com}. Each CV entry was segmented either into a set of sentences or long phrases using a various set of delimiters and special symbols used for list creation. In \cite{sayfullina_17}, you can find an example of CV from the same dataset. After segmenting each CV, we got 2517 text snippets in total.

We provided training and test data sizes in Table \ref{class_distr}.

\begin{table}
\begin{tabularx}{\textwidth}{s s s} 
	 & \textbf{Positive class} &  \textbf{Negative class} \\ 
    \hline
	\textbf{Job ads train} & 16,656 & 15,205  \\ 
    \hline
	\textbf{Job ads test} & 1984 & 222 \\ 
    \hline
	\textbf{Cv test} & 536 & 77 \\  
\end{tabularx} 
\caption{Class distribution in training and test datasets. Positive class refers to candidate soft skills. CV corpus was used only for testing. Both test datasets have naturally imbalanced class distribution. At the same time training data was intentionally made to be nearly balanced.}
\label{class_distr}
\end{table}

\section{Crowdsourcing skill annotations}\label{crowdsourcing}
In order to make a supervised dataset for training, we have established a crowd-sourcing experiment both for the job ads and for the CV corpora. We extracted all text snippets containing one soft skill from the list of 1\,072 soft skills. Each text snippet consists of the matched soft skill and a maximum of 10 words around a soft skill from both sides. Each snippet was shown with the soft skill highlighted in bold and a worker had to answer whether the highlighted skill is a soft skill referring to the candidate or not. 


A crowdsourcing experiment was done using CrowdFlower\footnote{https://www.crowdflower.com}, a crowdsourcing platform where workers can contribute to data annotation process. Each sentence was assessed by at least 3 workers. During the test run, the platform has selected only workers who got an accuracy of more than 85\% for the CV corpus (80\% for job ad corpus) on a limited labeled set of 40 sentences. The test run is made to select trusted workers for the main task. The rest 597 sentences were used for the main crowdsourcing experiment. The annotated sentences contained 149 unique skill phrases in total. All of the tagged CV data was used for the test experiment only. Obviously, some of the soft skills in the CV dataset were not seen during the training.


4863 sentences from the job ad corpus were labeled in a similar way as the CV corpus. Crowdsourcing experiment resulted in 427 skills labeled as non-candidate ones and 3984 as candidate ones. So we had to create more training data due to class imbalance. For that we picked all 9 soft skills that described non-candidate in more than 70\% of our annotations. Then we extracted 15,000 text snippets with those skills from the job ad corpus and marked them as non-candidate skills after discarding those containing ``candidate'', ``individual'' and ``looking for'' words near the soft skill. We also selected 123 skills that described only the candidates and mined 15,000 text snippets from the job ad corpus containing them and labeled these skills as candidate ones. The job ads test data contains only hand-annotated samples with 1984 positive (about a candidate) and 222 negatives labels. 

\section{Overview of the methods}\label{methods}

In this section we give an overview of three neural network classifiers used for our predictions and describe in detail various ways to represent a soft skill and its context for these classifiers.

\subsection{LSTM and CNN}

We have selected CNN\cite{cnn_sentence} and LSTM\cite{rumelhart1986learning,hochreiter1997long} models since they are both frequently used as text classification benchmark algorithms. 
In both models, words in the sentences are embedded with word2vec\cite{mikolov2013distributed} vectors. LSTM employs sequential nature of the text, handles long-term dependencies and allows making predictions on a variable length input. To make a prediction on variable length input, one can take the output of the hidden layer from the last word to make a final prediction. 

CNN model for text classification developed by Kim \cite{cnn_sentence} takes as input only a fixed-size 2D tensor, where rows correspond to words and columns correspond to word embedding dimensions. By applying convolutions with a window of N words covering full embedding dimension, CNN allows to take into account the word order, limited by the size of the window. We refer the reader to \cite{cnn_sentence} for more details about the model.

\subsection{Hierarchical Attention Model}

Hierarchical Attention Model (HAN) \cite{yang2016hierarchical} was shown to outperform CNN\cite{cnn_sentence}, LSTM, Conv-GRNN, LSTM-GRNN \cite{tang2015document} on six large classification benchmarks, including Yelp, IMDB, Amazon, Yahoo reviews.

The motivation of the model is the following: the output of classification depends more on certain words in certain sentences to which the most attention should be payed. Attention in the model is context-dependent, which means that similar words might be of different importance in different context. The model is called \textit{hierarchical} since it models documents using sentences which in their turn are modeled  sequentially using words. Gated Recurrent Units \cite{chung2015gated} with bidirectional structure are used to sequentially encode each sentence, as well as document as the sequence of sentences. 


\subsection{Proposed approaches for input representation}

In order to match candidate soft skills, it is important to consider the context in which the skill is detected. One baseline way to do this is to take the sentence where the skill is detected and train any text classification method to classify such sentences into those that refer to the candidate and those that refer to something else. However, a significant limitation of this approach is that the classifier is not informed about the location of the skill in the sentence. 

We propose three alternative ways of input representation for soft skill disambiguation:
\begin{enumerate}
\item Mask the soft skill phrase with \texttt{xxx} tokens for each word, in other words, apply \textit{soft skill masking}. E.g. the sentence ``bar and kitchen business seek a \textbf{dedicated person} who want to be'' will be transformed into ``bar and kitchen business seek a xxx xxx who want to be''.
\item Combine \textit{soft skill masking} with a soft skill embedding. 
\item Surround a soft skill phrase with \texttt{<begin>} and \texttt{<end>} tokens. We call this approach \textit{soft skill tagging}.
\end{enumerate}
While masking the soft skill phrase, we use the context around the skill, the position of the soft skill itself and the number of words in soft skill phrase. However, this approach does not employ for the prediction useful information about the skill itself, compared to two other approaches. In fact, each soft skill has a prior probability of describing a candidate and with soft skill masking this prior is not taken into account.

For the second alternative, we use both soft skill masking as well as soft skill embedding. Since the soft skill is hidden and not used while making the prediction, we came up with the idea of augmenting the classifier's input with a soft skill embedding. This way the prior information about the skill to make a prediction for is not lost, although the soft skill is still separated from its context. We suggest to augment the input to the last hidden layer of a neural network classifier with a soft skill embedding. Particularly, a neural network takes as a main input masked context, which goes through a forward pass. Then, an input to the last hidden layer is concatenated with a soft skill embedding and is used for a final prediction.The benefit of the this approach is that one can easily integrate it with any chosen neural network used in text classification. One approach to build a soft skill embedding is to use word2vec \cite{mikolov2013distributed} model and present text as the mean of its word-embeddings.


The third input representation would be a sentence with added \texttt{<begin>} and \texttt{<end>} tags around the soft skill in question. This approach is simple to implement, since it does not require any model changes as with soft skill embedding. Besides, the soft skill phrase is used for the classification, unlike with soft skill masking, and tagged as being ``important'' compared to raw sentence input from baseline approach. 

\section{Experimental evaluation}\label{results}

\subsection{Experimental settings}

All models except Hierarchical Attention were implemented from scratch using Python and PyTorch\cite{paszke2017automatic}. The code and the datasets will be publicly released here \footnote{\url{https://github.com/muzaluisa/soft-skill-matching}}.

Text preprocessing included case-folding, lemmatization with NLTK\cite{Loper:2002:NNL:1118108.1118117} WordNet lemmatizer as well as removing all syntax tokens, except commas.

We have set the width of CNN filters to be \{2,3,4\}. Since the network tends to overfit, we have used only 50 filters of each size. Maximum document length was set to 30 since we had a window of maximum 10 words around the soft skill from both sides. The  dropout \cite{dropout} rate was set to 0.5 and we found it to be useful for model regularization.  

LSTM network was implemented using variable-length inputs. The hidden layer size is 100, the dropout rate of 0.2 is applied to the input to the last fully connected layer to prevent over-fitting. We tested different dropout rates between 0.2 and 0.5 and found 0.2 to be the most optimal one. 

We have experimented with two types of word embeddings: Glove 6B pretrained word vectors on Wikipedia 2014 and Gigaword 5 datasets\footnote{\url{https://nlp.stanford.edu/projects/glove/}} and our own embeddings on full job ads data prepared using gensim library\cite{rehurek_lrec}. Finally, we chose own embeddings of size 100 for both CNN and LSTM. Furthermore, both networks used Adam optimizer with a learning rate of 0.001 due to its fast convergence and good performance. We used a batch size of 16 since often smaller batch sizes lead to better generalization \cite{keskar2016large}. Cross-entropy loss is used for learning by backpropagation. Half of the training data was used for validation of the model. 

Hierarchical Attention Model\footnote{The implementation of the model was adopted from: \url{https://github.com/EdGENetworks/anuvada}} has a GRU encoder with 50 hidden units, resulting in 100 units in both directions for sentence embedding. We left the majority of parameters of the implementation intact, including a batch size of 64, but used own pre-trained word vectors of size 100.

The training data consists of only job ads corpus, while for testing we used both job ads and CV corpora. The purpose of the CV experiment is to check how different approaches generalize on the new domain, where the distribution of the data is different. Since test data is imbalanced towards the positive class, we used F1-weighted measure. F1-weighted measure is a modification of F1-macro measure, but weighted with a class support. F1-macro measure computes precision and recall for each class separately and then takes the average with equal weight for each class.

The performance of compared approaches is evaluated based on the precision-recall balance. While using naive approach by selecting all soft skill phrases we get 89.50\% precision on the job corpus and 87.5\% precision on the CV corpus with 100\% recall. Since the aim is to increase the precision while soft skill matching with still reasonable recall, we fixed a precision to be ~95\% and ~90\% for the job and CV corpora respectively by adjusting a decision boundary between two classes. A lower precision for the CV corpus was chosen, since it is not used during the training and therefore it is more difficult to achieve 95\% of precision with a reasonable recall.

\begin{table}[!htp]
\caption{F1-weighted score, precision and recall, (\%) for compared classifiers across four different input representations. For comparison, we fixed the precision for job corpus to be ~95\% and ~90\% for CV corpus by adjusting the decision boundary between two classes. Naive approach means labeling all matched skills as candidate soft skills and F1-weighted score is not applicable. Soft skill tagging showed the highest recall on the test samples from the job corpus used for the main evaluation.}
\begin{tabularx}{\textwidth}{b s s s | s s s}
\\
 & \multicolumn{3}{c}{Job test} & \multicolumn{3}{c}{CV test} \\
\hline 
\textbf{Method} & \textbf{F1} & \textbf{Pr} & \textbf{Re} & \textbf{F1} & \textbf{Pr} & \textbf{Re} \\
\hline 
Naive approach & N/A  & 89.50 & 100 & N/A & 87.4 & 100 \\
\hline
\textbf{Unmodified sentence} \\
\hline
CNN & 80.07 & 95.5 & 76.69 & \textbf{82.23} & \textbf{90.26} & \textbf{88.80} \\
 LSTM & 82.81 & 95.03 & 81.69 & \textbf{82.20} & \textbf{90.24} & \textbf{88.80} \\
HAN & 77.85 & 95.35 & 73.29 & 78.14 & 90.68 & 80.00 \\
\hline
\textbf{Soft skill masking} \\
\hline 
CNN & 79.44 & 95.19 & 76.02 & 79.92 & 90.36 & 84.00 \\
LSTM & 78.95 & 95.06 & 75.34 & 78.79 & 90.79 & 81.12 \\
HAN & 82.8 & 95.07 & 81.7 & 47.04 & 90.34 & 34.95 \\
\hline
\multicolumn{3}{l}{\textbf{Soft skill masking and embedding}} \\
\hline
CNN & 78.17 & 95.30 & 73.80 & 75.08 & 90.14 & 75.18 \\
LSTM & 82.08 & 95.06 & 80.40 & 76.632 & 90.52 & 77.46 \\
HAN & 79.87 & 95.01 & 76.81 & 44.75 & 91.05 & 32.34 \\
\hline
\textbf{Soft skill tagging} \\
\hline
CNN & 80.36 & 95.13 & 77.54 & 80.9 & 90.23 & 86.19 \\
LSTM & \textbf{84.18} & \textbf{95.05} & \textbf{83.92} & 81.33 & 90.11 & 87.31 \\
HAN & 81.28 & 95.21 & 78.94 & 90.07 & 89.91 & 82.56 \\
\hline
\end{tabularx}
\label{tab:results}
\end{table}

\subsection{Baseline approach}

For the baseline approach we used unmodified sentences without replacing a soft skill. This way, it is not clear from the representation for which soft skill the prediction is made. At the same time, many sentences contain only one skill that can be given higher weight for the prediction without additional tagging of the skill. Thus, the classifier could learn soft skills from the training data as well as features, indicating whether the sentence will contain candidate soft skills. 

We combined the results of all our experiments in Table \ref{tab:results}. 
Surprisingly, but from Table \ref{tab:results} we can see that \textit{unmodified sentence} representation gives the best result on the CV corpus. CNN and LSTM using this representation result in almost the same performance on the CV corpus and we highlight both methods as the strongest ones. 

\subsection{Soft skill masking}

Masking a skill (with and without soft skill embedding) from the sentence turned out to be overall the weakest from the proposed representations. At the same time the sentence a with hidden or masked skill have a reasonable performance, meaning that the context itself can be used for the prediction. For CNN and LSTM soft skill masking did not bring any improvement compared to the raw sentence representation.

The third input representation used additionally the mean of word2vec \cite{mikolov2013distributed} embedding computed from the soft-skill phrase. Since the number of added dimensions for both CNN and RNN is almost (90 for CNN, 100 for LSTM) the size of the last hidden layer, the prior coming from the soft skill is strong. Surprisingly, but the embedding resulted in worse performance than with the soft skill masking. There can be several reasons for that including the lack of various soft skill samples for learning from their embeddings as well as overfitting due to a strong soft skill prior. 

\subsection{Soft skill tagging}

Soft skill tagging similarly to an unmodified sentence representation keeps the soft skill inside its context. These two approaches turned out to show overall the highest performance. From Table \ref{tab:results} we can see that a soft skill tagging using LSTM gives the highest recall of 83.92\% on the job corpus across all methods and representations. Besides, the performance of CNN on the test samples from the job corpus is also the highest with soft skill tagging representation.  

In addition, HAN model did not outperform other models in our task across various representations. This could happen because our inputs consisted of only one sentence, while a HAN model assumes input samples to consist of several sentences. Besides, the recall on the CV corpus was pretty low while using soft skill masking.

\subsection{Skill disambiguation use case}

\begin{figure}[!tph]
     \centering
     \includegraphics[scale=0.5]{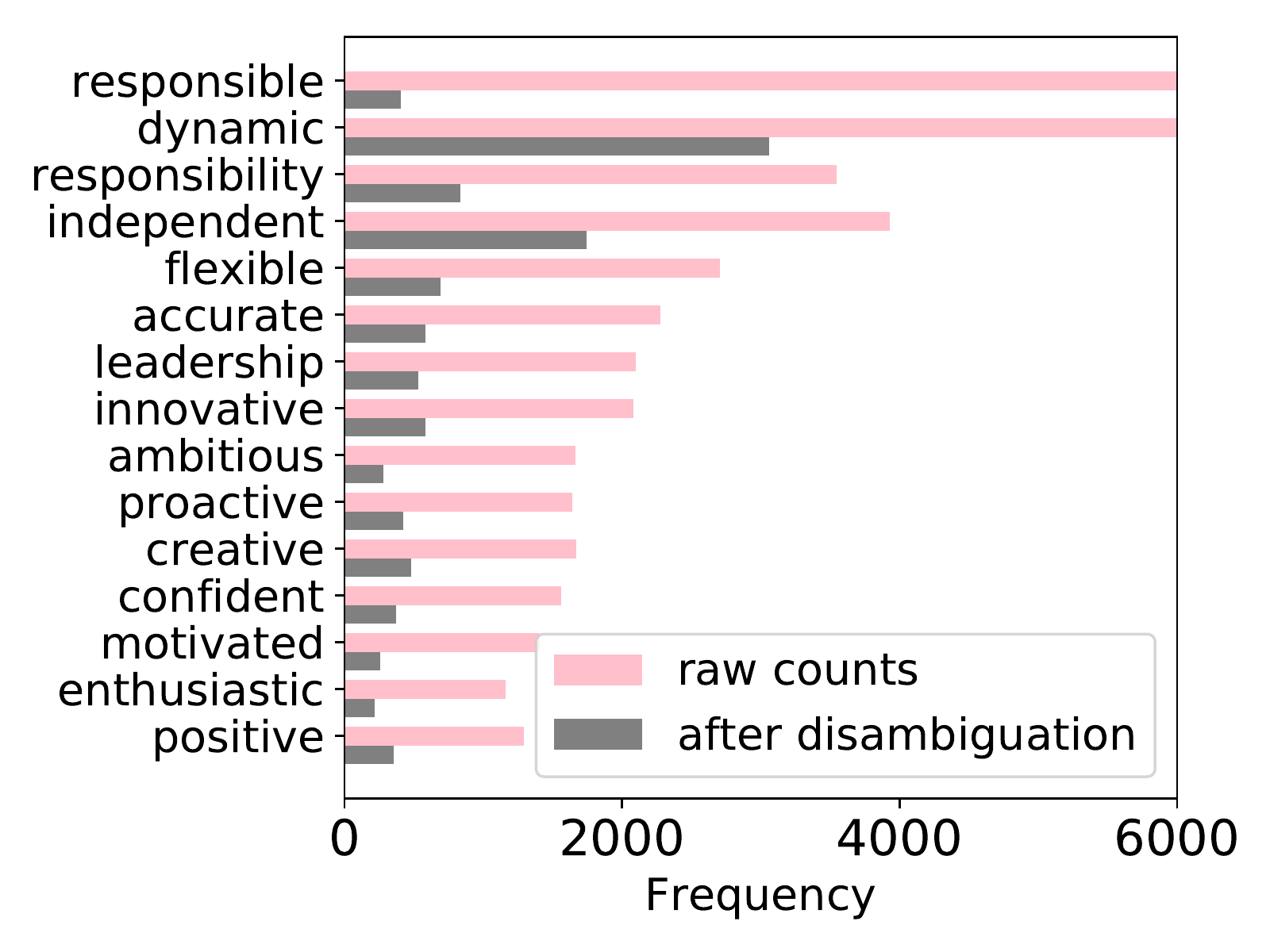}
   \caption{Raw counts in pink correspond to raw frequencies, while the bars in gray correspond to counts after filtering out skills not referring to the candidate. We showed the skills for which the difference between these counts was the largest to show the amount of possible false positive matches for some skill phrases with a naive approach.}
     \label{fig:before_after}
 \end{figure}
 
We decided to show how drastically soft skill frequencies might change
after we filter out words not related to the candidate.
For that we took randomly 100,000 job ads with soft skills of maximum length equal to two since most ambiguous phrases tend to be short. Then we ran disambiguation classifier based on the LSTM and soft skill tagging and filtered out the skills with a high confidence score of not referring to the candidate.
Skills with top difference are given in Figure~\ref{fig:before_after}.
Moreover, there are 58 skills that had more than 50 words filtered out. These results show the amount of possible false positive results for certain skills while using a naive matching giving raw counts. 

 


\section{Conclusion}\label{conclusions}

Soft skill analysis and automatic detection of soft skills in unstructured text has not been systematically approached previously. A naive detection of soft skills in unstructured text leads to false positive matches, when a skill does not refer to the candidate. 
To tackle this problem, our work proposes a method for differentiating between soft skills referring to a candidate and other skills using binary classification. The prediction is made using the context where the soft skill occurs. We propose several approaches for making an input representation to the classifier, including raw context sentences, context sentences with soft skill masking, soft skill tagging and augmentation with soft skill embedding. 

We compare several strong neural network baselines for text classification and show that LSTM yields the best disambiguation performance on sentences with tagged soft skills on the job corpus and with a raw sentence representation on the CV corpus. 
The developed soft skill disambiguation methods open up new possibilities for refined analyses of soft skill demands and supply on the job market.

\bibliographystyle{splncs03.bst}
\bibliography{bibl}

\end{document}